\documentclass[letterpaper]{article} 
\usepackage{aaai24}  
\usepackage{times}  
\usepackage{helvet}  
\usepackage{courier}  
\usepackage[hyphens]{url}  
\usepackage{graphicx} 
\usepackage{natbib}  
\usepackage{caption} 
\usepackage{algorithm}
\usepackage{algorithmic}

\usepackage{newfloat}
\usepackage{listings}
\DeclareCaptionStyle{ruled}{labelfont=normalfont,labelsep=colon,strut=off} 
\floatstyle{ruled}
\newfloat{listing}{tb}{lst}{}
\floatname{listing}{Listing}
\usepackage{amsmath,amsfonts,amssymb}
\usepackage[caption=false,font=normalsize,labelfont=sf,textfont=sf]{subfig}
\usepackage{textcomp}
\usepackage{verbatim}
\usepackage{cite}
\usepackage{multirow}
\usepackage{makecell}
\usepackage{booktabs}

\newcommand \footnoteONLYtext[1]
{
 \let \mybackup \thefootnote
 \let \thefootnote \relax
 \footnotetext{#1}
 \let \thefootnote \mybackup
 \let \mybackup \imareallyundefinedcommand
}

\title{Learning to Manipulate Artistic Images}
\author
{
    Wei Guo\equalcontrib,
    Yuqi Zhang\equalcontrib,
    De Ma\corrcontrib,
    Qian Zheng\corrcontrib
}
\affiliations
{
    Zhejiang University \\
    \{snailforce, yq\_zhang, made, qianzheng\}@zju.edu.cn
}

\usepackage{bibentry}

\begin{document}

\maketitle

\begin{abstract}
Recent advancement in computer vision has significantly lowered the barriers to artistic creation. 
Exemplar-based image translation methods have attracted much attention due to flexibility and controllability. 
However, these methods hold assumptions regarding semantics or require semantic information as the input, while accurate semantics is not easy to obtain in artistic images. 
Besides, these methods suffer from cross-domain artifacts due to training data prior and generate imprecise structure due to feature compression in the spatial domain. 
In this paper, we propose an arbitrary Style Image Manipulation Network (SIM-Net), which leverages semantic-free information as guidance and a \textit{region transportation} strategy in a self-supervised manner for image generation. 
Our method balances computational efficiency and high resolution to a certain extent. 
Moreover, our method facilitates zero-shot style image manipulation. 
Both qualitative and quantitative experiments demonstrate the superiority of our method over state-of-the-art methods.
Code is available at https://github.com/SnailForce/SIM-Net.
\end{abstract}
\section{Introduction}

Art can cultivate sentiment, improve self-cultivation, and inherit culture.
The use of artificial intelligence in the process of creating art was significantly accelerated with rapid advances in machine learning. 
Artificial Intelligence (AI) has endowed art with more possibilities in various artistic fields, such as calligraphy, imagery, design, and others, thereby bridging the gap between people and art.
Particularly, image manipulation has unique creativity and strong interactivity, thereby further lowering the barrier to the artistic creation.

\begin{figure}[t]
    \centering
    \includegraphics[width=0.48\textwidth]{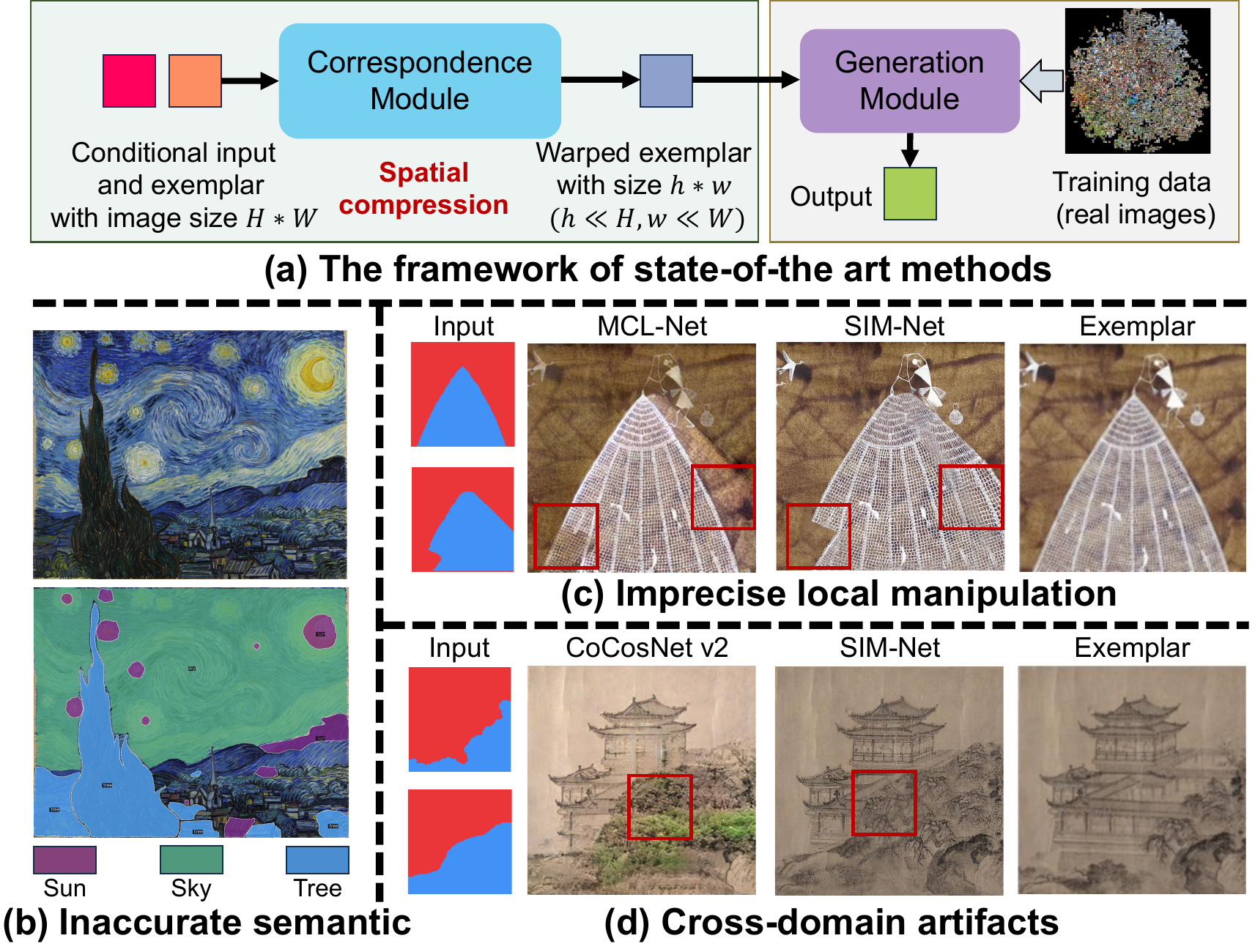}
    \caption
    {
    (a) The framework of state-of-the-art exemplar-based image translation methods, such as CoCosNet v2 \cite{zhou2021cocosnet}, MCL-Net \cite{zhan2022marginal}, and MATEBIT \cite{jiang2023masked}.
    (b) These methods require accurate semantic conditional input, while accurate semantic information of artistic images is difficult to extract.
    (c) The spatial compression in the cross-domain alignment phase leads to imprecise local details.
    (d) The conditional generation phase might introduce cross-domain artifacts.
    }
    \label{fig:head}
\end{figure}

Image manipulation essentially falls into the category of image-to-image translation, which has demonstrated success across a wide range of applications \cite{isola2017image, liu2017unsupervised}. 
Different from other conditional image-to-image translation methods, exemplar-based image translation enables more flexible user control and better generation quality, which especially is well-suited for manipulating artistic images.
These methods \cite{zhang2020cross, zhou2021cocosnet, zhan2022marginal, zhan2021unbalanced, zhang2022easypainter} consist of two main phases: cross-domain alignment (align exemplar and conditional input to get aligned representation) and conditional generation (Figure \ref{fig:head}(a)).

However, such methods have following problems when applied to artistic images.
\textbf{Semantic input.} 
Existing methods are either based on specific semantic scenarios (\textit{e.g.}, face \cite{liu2015deep, fan2022facial}, human pose \cite{ma2017pose, men2020controllable, zhang2021deep}, etc.) or require semantic labels \cite{zhang2020cross, zhou2021cocosnet, zhang2023adding}. 
However, extracting accurate semantic information from artistic images as input is challenging (Figure \ref{fig:head}(b)).
\textbf{Imprecise control.}
Image manipulation requires fine-grained control, while feature compression like multi-level feature pyramids \cite{zhou2021cocosnet, jiang2023masked} leads to imprecise local details (Figure \ref{fig:head}(c)).
\textbf{Computational efficiency.}
Directly generating high-resolution images brings memory overhead at a square level in terms of image size, making it unfeasible. 
Some methods \cite{zhan2022marginal, jiang2023masked} utilize feature pyramids to generate high-resolution images by employing low-resolution images as guidance. 
However, these methods still exhibit significant computational overhead.
\textbf{Cross-domain artifacts output.} 
Existing methods generate images based on the correspondence and the aligned representation, which is typically implemented by unsupervised generative models such as GANs or Diffusion models. 
However, the correspondence is not always accurate.
For inaccurate regions, these methods generate images according to prior knowledge learned from training data which will cause cross-domain artifacts (Figure \ref{fig:head}(d)).
The style of testing data and training data in artistic image manipulation can hardly be guaranteed to be the same or similar, making it challenging to utilize data priors for generation.

In this paper, we propose an arbitrary \textbf{S}tyle \textbf{I}mage \textbf{M}anipulation \textbf{N}etwork (SIM-Net) for exemplar-based artistic image translation, which consists of Mask-based Correspondence Network and Translation Network, similar to the mainstream methods (Figure \ref{fig:head}(a)).
The Mask-based Correspondence Network takes low level information as input due to the difficulty of extracting semantic information from artistic images, and we use semantic-free masks which have better regional control capability \cite{he2017mask}.
Afterwards, we utilize the Mask-based Correspondence Network to establish the correspondence between two masks guided by the exemplar. 
These two masks exhibit substantial overlapping regions and few non-overlapping regions.
Therefore, we utilize a few number of keypoints to adaptively control different regions for local alignment with low computational overhead implemented by the Local Region Alignment Module.
Subsequently, to further obtain the full-resolution correspondence, we dilate these local regions into the global image space and obtain full-resolution warp fields implemented by the Dilating Module.
While these several warp fields are full resolution, they focus on respective keypoints regions and provide more precise control. 
Therefore, we utilize the Generation Network to merge well-controlled regions corresponding to several warp fields to generate precise images. 
For overlapping regions between masks, we consider using the original exemplar as background estimation to participate in the merging process.
Consequently, we propose a \textit{region transportation} strategy to generate images in a self-supervised manner instead of an unsupervised manner implemented by the Image Transport Module, thereby avoiding introducing style features from other domains and cross-domain artifacts.
However, this approach introduces splicing artifacts between local regions.
To eliminate splicing artifacts, we construct pseudo ground truth to provide both geometrically consistent and spatially consistent supervisory signals implemented by the Texture-Guidance Module, and propose a style self-supervised strategy for training. 
Our contribution can be summarized as follows:
\begin{itemize}
    \item We propose a zero-shot arbitrary Style Image Manipulation Network SIM-Net, which does not need to touch any style training data and effectively eliminates cross-domain artifacts.
    \item We propose a Mask-based Correspondence Network, which ensures a balance between computational efficiency and high resolution, and a \textit{region transportation} strategy for generation in a self-supervised manner.
    \item Experimental results demonstrate that our method outperforms previous state-of-the-art methods in terms of Style Loss, SSIM, LPIPS, and PSNR.
\end{itemize}

\section{Related Work}

\subsection{Exemplar-Based Image Translation}

Early pioneering works \cite{zhu2017unpaired, park2019semantic, xu2020generative} attempt to achieve global control over style consistency in generated images by extracting latent codes using style encoder. 
However, these methods neglected spatial correlations between an input image and an exemplar image, thus failing to produce precise local details.
Recently, exemplar-based image translation methods \cite{zhang2020cross, zhou2021cocosnet, zhan2021unbalanced, zhan2022marginal, jiang2023masked}, which leverage an exemplar image to control the style of translated images, establish dense correspondence, and warp the exemplar for generation, have attracted increasing attention.
However, as illustrated in Figure \ref{fig:head}, these methods are difficult to obtain accurate semantic information in artistic images, and suffer from imprecise local control, substantial computational overhead, and cross-domain artifacts.
In contrast, we address these problems.

\subsection{Low Level Information Guided Methods}

The effectiveness of low level priors has been demonstrated in several vision tasks, such as image super-resolution \cite{li2023learning}, image inpainting \cite{lugmayr2022repaint}, and image restoration \cite{dogan2019exemplar}. 
Existing works \cite{nazeri2019edgeconnect, zhang2023adding} use low level information such as canny edge and sketch for generation.
It is worth noting that some of these methods focus on specific semantic scenarios (face \cite{chen2020deepfacedrawing}, human pose \cite{li2023cee}, dog \cite{pan2023drag}, etc.).
Despite their input belonging to low level information, we do not discuss.
In contrast to tasks addressed by these methods, we utilize low level information for manipulating artistic images, avoiding cross-domain artifacts associated with high level smenatic information.

\subsection{Motion Transfer}

The motion transfer task aims to drive human body motion based on the given video.
These methods \cite{siarohin2019first, siarohin2021motion} achieve smooth and natural movement of the human body by capturing the local affine transformations between video frames and a single image to establish global correspondence, which essentially estimates dense optical flow fields between video frames.
We refer to the idea of these methods and regard the problem of artistic image manipulation as a mask motion problem.

\subsection{Artificial Intelligence Based Art Research}

Visual art research is a mysterious and remote discipline. 
Existing work focuses on problems of painting classification \cite{ma2017part}, style identification \cite{chen2019recognizing}, style transfer \cite{xu2023learning}, painting caption \cite{bai2021explain}, and photo stylization \cite{he2018chipgan}.
Different from existing research, we focus on manipulating artistic images of arbitrary style.

\section{Proposed Method}
\begin{figure*}[!ht]
    \centering
    \includegraphics[width=0.92\textwidth]{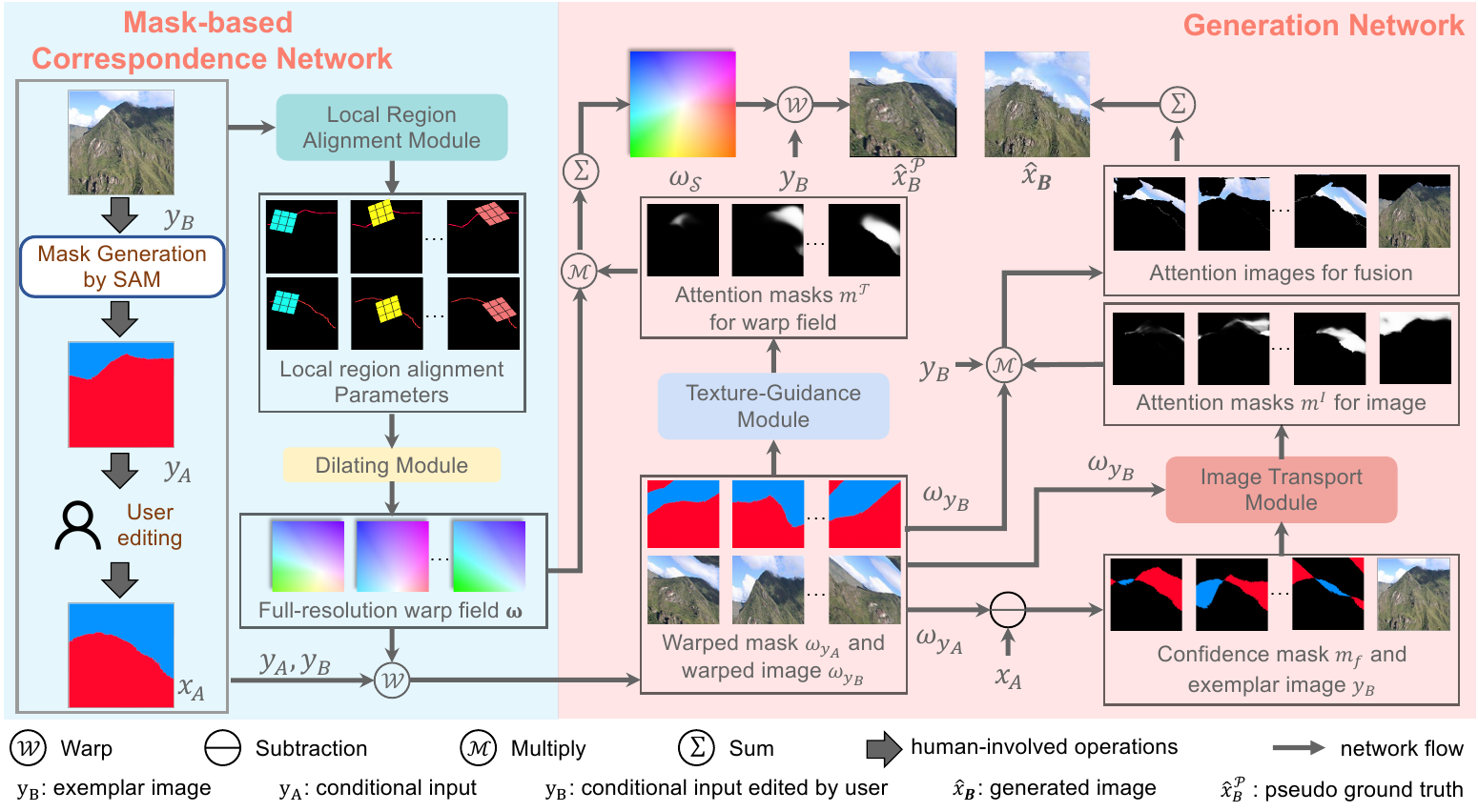}
    \caption{
    The overall architecture of SIM-Net. 
    The SAM module is used to extract the semantic-free mask of exemplar, denoted as $y_A$, which is then edited by users to obtain the conditional mask, denoted as $x_A$.
    First, the Local Region Alignment Module is used to generate a few number of keypoints that adaptively govern modified regions.
    Subsequently, the Dilating Module is employed to establish multiple full-resolution corresponding warp fields corresponding to keypoints for global control.
    Notably, these wrap fields exhibit better control over the region near their corresponding keypoints.
    Finally, to utilize the characteristics of warp fields, we propose the \textit{region transportation} strategy implemented by the Image Transport Module, utilizing multiple warp fields to construct the generated image, denoted as $\hat{x}_B$.
    However, $\hat{x}_B$ exhibits splicing artifacts marked by spatial inconsistency.
    We further design the Texture-Guidance Module to construct the pseudo ground truth, denoted as $x_B$, serving as a self-supervised signal to eliminate splicing artifacts to ensure spatial consistency.
    }
    \label{fig:framework}
\end{figure*}

Given an input image $x_A\in\mathbb{R}^{H\times W}$ in domain $\mathcal{A}$ and an exemplar artistic image $y_B\in\mathbb{R}^{H\times W\times 3}$ in domain $\mathcal{B}$, our goal is to generate a image $x_B\in\mathbb{R}^{H\times W\times 3}$ which preserves low level information in $x_A$ and style information in $y_B$. 
The exemplar $y_B$ can have arbitrary styles.
Figure \ref{fig:framework} illustrates an overview of our framework SIM-Net.
First, we use semantic-free masks to avoid introducing inaccurate semantic information.
Second, we utilize a few number of keypoints to generate full-resolution warp fields, which ensures a balance between computational efficiency and high resolution.
Third, we propose a \textit{region transportation} strategy for image generation to avoid cross-domain artifacts.

\subsection{Mask-Based Correspondence Network}
\label{sec:eman}

\textbf{Low Level Input.} 
Previous works \cite{zhan2022marginal, jiang2023masked} use an exemplar with a semantic mask to build correspondence. 
However, the semantic information of artistic images is inaccurate, so we consider using masks with strong region control ability without using semantic information. 
Segment anything exactly can provide a semantic-free mask $y_A\in \mathbb{R}^{H\times W}$. 
Users can get $x_A$ by editing the input image $y_A$, expressed as a binary mask.
We consider aligning two masks guided by the exemplar and formulate the problem as:
$\mathcal{M} (x_A, y_A | y_B)$.

\noindent\textbf{Local Region Alignment Module.} 
Previous methods use techniques such as feature pyramids \cite{zhou2021cocosnet, liang2021high} to build correspondence which are computationally substantial, and lack precise local control. 
In artistic image manipulation, two masks primarily maintain unchanged across most regions, with alteration occurring only in specific modified regions ($x_A$ and $y_A$ in Figure \ref{fig:framework}). 
Therefore, we propose a strategy that utilizes a few number of keypoints to adaptively govern the modified regions, thereby achieving local alignment.
Notably, keypoints have lower computational overhead.

The local alignment between these keypoints is achieved through affine transformation, which is essentially a piecewise linear approximation of the contour so that the original contour becomes the target contour.
As the contours are gradually aligned, the corresponding regions are adaptively aligned.
In particular, the affine transformation between two masks is essentially a problem of isomorphic change.

We follow the \cite{siarohin2019first} which is based on a set of learned keypoints with local affine transformations for video sequence tasks to support complex motions, and trained in an unsupervised manner.
In our case, we build upon the concept of keypoints to achieve local alignment with lower computational overhead.
Multiply $y_A$ and $y_B$ with $x_A$ as input, we can get keypoints $p_1, \cdots, p_K$ and $K$ heatmaps $H^1, \cdots , H^K, s.t.H^k\in [0,1]^{H\times W}$, and further calculate the affine transformation parameters.
Each group of parameters represents image transformation, representing the affine transformation from the virtual reference image $R$ to the input image, specifically formulated as $A_{input \leftarrow R}^k$.

\noindent\textbf{Dilating Module.} 
Although the above module can effectively reduce the computational overhead, we still need global correspondence for accurate manipulation control.
Different from the previous method \cite{siarohin2019first, siarohin2021motion}, which only focuses on local regions, we consider dilating these local regions into the global image space to obtain full-resolution correspondence, denoted as warp fields $\omega$.
The implementation is similar to \cite{siarohin2019first} and under our task can be deduced as follows:

\begin{align}
    \omega^k &= \mathcal{D} (x_A, y_A, y_B, A^k)
    = A_{x_A\leftarrow R}^k
        \begin{bmatrix}
            A_{y_A'\rightarrow R}^k \\
            0~~0~~1
        \end{bmatrix}^{-1}, \\
    y' &= y_A\otimes  y_B,
\end{align}
where $\mathcal{D}$ presents the Dilating Module and $y'$ denotes the combination of $y_A$ and $y_B$.

As illustrated in Figure \ref{fig:framework}, $\omega$ is then applied to $y_A$ and $y_B$, generating the warpped image $\omega_{y_B}$ and the warpped mask $\omega_{y_A}$. 
In addition, the full-resolution warp field can be used as a global precise control signal to provide supervisory information for subsequent operations.
Through the combination of the above modules, we guarantee the trade-off between computational overhead and full resolution.

\subsection{Translation Network}
Existing methods are trained on realistic images. 
When dealing with artistic image manipulation, these methods generate images based on the training data prior, introducing cross-domain artifacts.
Therefore, we utilize \textit{region transportation} for image generation.
Specifically, we propose the Image Transport Module, which can ensure strict control over segmentation and avoid introducing additional content by \textit{region transportation} based on the warp fields.
However, \textit{region transportation} will bring splicing artifacts between regions.
Furthermore, we propose the Texture-Guidance Module to eliminate artifacts.
Details are described below.

\noindent\textbf{Image Transport Module.} 
The previous module yields $K$ sets of full-resolution warp fields, which are derived through keypoint-based local transformations.
A single warp field provides more precise control within its own regions, whereas other warp fields provide more control within their respective regions. 
Therefore, we consider merging them to obtain a more accurate result. 
We observe that the majority of the two masks remain unchanged and the control capability of these warp fields for unchanged regions is limited.
We further incorporate the original exemplar image for background estimation to participate.

Specifically, we propose a strategy to filter out regions with low confidence in each warp field and merge high confidence regions to achieve global control by weighted fusion of attention mechanism.
Subsequently, we construct a confidence mask used for filtering out regions, denoted as $m_f$:

\begin{equation}
    \{m_f^i; y_B\}=\{x_A-\omega_{y_A}^i, i=1, \cdots, K;y_B\}.
\end{equation}

To achieve more effective control, we compell $y_A$ towards $x_A$, and strictly ensure the close fit of local contours in a self-supervised manner.
Especially, we use $K$ warped masks $\omega_{y_A}$, the original unwarped image for background estimation and $K$ filter masks, and output $K+1$ fused attention maps $m_i^I\in \mathbb{R}^{H\times W}, i=0,\ldots, K$, which are activated based on softmax to ensure that the sum at each pixel is 1. 
The original unwarped image is used to keep the content of the unmodified region, which is far from the keypoints:

\begin{equation}
    m_i^I(p)=\frac{exp(m_i^I(p))}{\sum_{i=0}^{K}exp(m_i^I(p))}, i=0,\ldots, K,
\end{equation}
where $m_i^I(p)$ represents the value of $m_i^I$ at the space coordinate position p. 
Finally, the image with segmented control is obtained through weighted fusion:

\begin{equation}
    \label{equ:fuse}
    \begin{split}
        \hat{x}_A = \sum_{i=0}^{K}m_i^I\cdot \omega_k(y_A),\\
        \hat{x}_B = \sum_{i=0}^{K}m_i^I\cdot \omega_k(y_B).
    \end{split}
\end{equation}

As $x_A$ and $y_A$ are paired masks, which provide supervised guidance, we incorporate the BoundaryIoU Loss \cite{cheng2021per} $L_{\rm {bound}}$ to minimize the difference between the warped masks $\omega_{y_A}$ and $x_A$, which can get better confidence masks and facilitate the training of the Mask-based Correspondence Network:

\begin{equation}
    \mathcal{L}_{\rm bound}=1-{\rm BoundaryIoU}(\omega_{y_A}^i, x_A), i=1,\cdots, K.
\end{equation}

\noindent\textbf{Texture-Guidance Module.} 
The Image Transport Module brings splicing artifacts between local regions as illustrated by $\hat{x}_B$ in Figure \ref{fig:styles}.
We need ground truth with spatial consistency as supervision, that is, the relative position is determined without splicing artifacts.
Subsequently, a trival way is to use the exemplar as ground truth.
However, the exemplar is geometrically inconsistent with $\hat{x}_B$, that is, the geometry layout is different, and we lack a means to quantitatively evaluate the similarity between them to facilitate training.
Therefore, we further construct a pseudo ground truth that is both spatially consistent and geometrically consistent.
Considering warp field $\omega$ based on affine transformation parameters, which exhibits strong geometric consistency and spatial consistency, we propose to obtain pseudo ground truth by performing a \textbf{single warp} on the original exemplar.

Compared with the fusion of the previous module in the image space, we choose to fuse in the warp field space to obtain a single warp field $\omega_\mathcal{S}$ by a similar attention module:

\begin{figure}[t]
    \centering
    \includegraphics[width=0.48\textwidth]{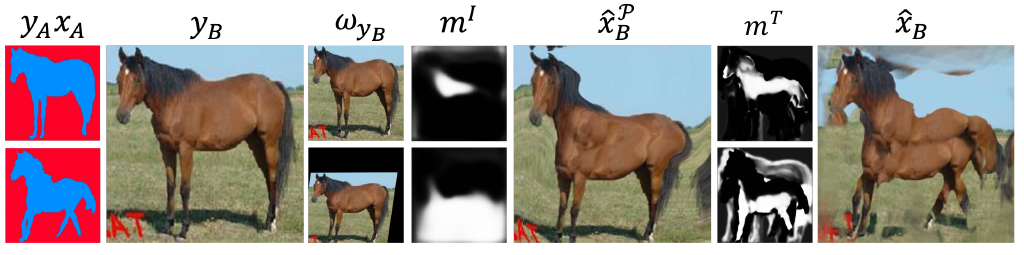}
    \caption
    {
        The intermediate results of a training sample in the early epoch. 
        It is evident that $\hat{x}_B^{\mathcal{P}}$ exhibits better geometric consistency and spatial consistency in the early epoch, thanks to the geometric consensus achieved through the warp fields. 
        However, the layout of $\hat{x}_B^{\mathcal{P}}$ is less consistent with $x_A$.
        Additionally, it can be observed that $\hat{x}_B$ demonstrates improved semantic consistency as a result of the fusion of several candidates. 
        However, this process can introduce fusion and splicing artifacts that may disrupt the geometric consistency and spatial consistency.
        The results provide visual evidence of the trade-off between geometric consistency, spatial consistency, and semantic consistency in the intermediate results during the early training epochs.
    }
    \label{fig:styles}
\end{figure}

\begin{align}
    \omega_\mathcal{S}(p) &= \sum_{i=1}^{K}m_i^T(p)\cdot \omega_i(p), \\
    m_i^T(p)&=\frac{exp(m_i^T(p))}{\sum_{i=0}^{K}exp(m_i^T(p))}, i=1, \ldots, K,
\end{align}
where $m_i^T(p)$ represents the value of $m_i^T$ at the space coordinate position p.


The pseudo ground truth $\hat{x}_A^{\mathcal{P}}$ and $\hat{x}_B^{\mathcal{P}}$ are obtained by swaping $y_A$ and $y_B$ through a single warp field $\omega_\mathcal{S}$.
We further propose the style self-supervision strategy, which constrains the output $\hat{x}_B$ to maintain style consistency with the pseudo ground truth $\hat{x}_B^{\mathcal{P}}$, as measured by the Style Loss \cite{gatys2016image}.
We use Contextual Loss \cite{mechrez2018contextual} to minimize the style difference. 
The contextual loss is computed by:

\begin{equation}
    \label{equ:cl}
    \mathcal{L}_{{\rm context}} = \sum_l w_l[-log(CX(\phi^l(\hat{x}_B), \hat{x}_B^{\mathcal{P}}))],
\end{equation}
where $w$ is used to adjust the weights between different network layers to balance the loss of each layer, $CX(x, y)$ represents the similarity between feature maps.

Compared to MSE Loss and SSIM Loss, which have stricter structural constraints, our method, as mentioned before, ensures structural consistency, and can more effectively enforce constraints on color and texture. 
In addition, compared with directly using the exemplar $y_B$ as style information guidance, such a strategy can further reduce the impact of image semantic content on style feature extraction, and pay more attention to the differences brought by style.

To facilitate training and ensure the generation of images with geometric consistency and spatial consistency, we use the cycle loss function $\mathcal{L}_\text{cyc}$.
Specifically, through swapping $x_A$ and $y_A$, we utilize $\hat{x}_B^{\mathcal{P}}$ as the exemplar and generates the pseudo original image $\hat{y}_B$.
The cycle loss function minimizes the $\ell_1$ distance between $y_B$ and $\hat{y}_B$.
It is worth noting that both the Image Transport Module and Texture-Guidance Module are trained based on the same Mask-based Correspondence Network, thereby constraining each other.

\subsection{Additional Loss Functions}
We use Equivariance constraint Loss $\mathcal{L}_{\rm {eq}}$ \cite{siarohin2019first} to ensure the stable training of the Local Region Alignment Module, use Perceptual Loss $\mathcal{L}_{\rm {perc}}$ to constrain the image structure, and use Conditional alignment Loss $\mathcal{L}_{\rm {mask}}^I$ and $\mathcal{L}_{\rm {mask}}^T$ to ensure consistency of mask structure.

\noindent\textbf{Total Loss.}
Our framework is end-to-end optimized to jointly achieve zero shot arbitrary style image manipulation and alternate between optimizing Mask-based Correspondence Network, Image Transport Module, and Texture-Guidance Module using the aforementioned loss functions.
The overall objective function of proposed framework is:

\begin{equation}
    \begin{split}
        \mathcal{L}_{\rm {total}} = &\lambda_1 \mathcal{L}_{\rm {eq}} + \lambda_2 \mathcal{L}_{\rm {perc}} + \lambda_3 \mathcal{L}_{\rm {context}} + \lambda_4 \mathcal{L}_{\rm {bound}}\\
    &+  \lambda_5 (\mathcal{L}_{\rm {mask}}^I + \mathcal{L}_{\rm {mask}}^S) + \lambda_6 \mathcal{L}_{\rm {rec}} + \lambda_7 \mathcal{L}_{\rm {cyc}},
    \end{split}
\end{equation}
where $\lambda$ is weighting parameter.

\noindent\textbf{Inference.} 
The output of $\hat{x}_B$ is taken as the manipulation result.
Note that manipulating several regions could be achieved by sequentially feeding the output to the model. 

\section{Experiments}
\label{Experiments}

\noindent\textbf{Implementation details.}
The learning rate for the framework is $2e-4$, and we use the Adam solver with $\beta_1 = 0.5$ and $\beta_2 = 0.999$ for the optimization. 
$K$ is set to be 10.
Unless otherwise specified, the resolution of generated images for translation tasks is $256\times 256$ for fair comparison.
The experiments are conducted using 4 TITAN Xp GPUs.


\begin{table*}[t]
    \setlength\tabcolsep{4pt}
    \centering
    \resizebox{\textwidth}{!}
    {
        \begin{tabular}{l|ccc|ccc|ccc|ccccc}
        \bottomrule
        & \multicolumn{3}{c|}{\scriptsize \textbf{{\textsc{Ab}}}, \textbf{{\textsc{Cu}}}, \textbf{{\textsc{In}}} (73)}
        & \multicolumn{3}{c|}{\scriptsize \textbf{ {\textsc{Ch}}, {\textsc{Ex}}, {\textsc{Ja}}, {\textsc{Mo}}} (95)}
        & \multicolumn{3}{c|}{\scriptsize \textbf{{\textsc{Im}}, {\textsc{Ph}}, {\textsc{Su}}} (69)}
        & \multicolumn{5}{c}{\scriptsize \textbf{All} (237)} \\
        \cmidrule { 2 - 15 }
        & \makecell[c]{Style Loss $\downarrow$ \\(ROI)} & \makecell[c]{SSIM $\uparrow$ \\ (ROU)} & \makecell[c]{Style Loss $\downarrow$ \\(Whole)}
        & \makecell[c]{Style Loss $\downarrow$ \\(ROI)} & \makecell[c]{SSIM $\uparrow$ \\ (ROU)} & \makecell[c]{Style Loss $\downarrow$ \\(Whole)}
        & \makecell[c]{Style Loss $\downarrow$ \\(ROI)} & \makecell[c]{SSIM $\uparrow$ \\ (ROU)} & \makecell[c]{Style Loss $\downarrow$ \\(Whole)}
        & \makecell[c]{Style Loss $\downarrow$ \\(ROI)} & \makecell[c]{SSIM $\uparrow$ \\ (ROU)} & \makecell[c]{Style Loss $\downarrow$ \\(Whole)} & \makecell[c]{PSNR $\uparrow$ \\ (ROU)} & \makecell[c]{LPIPS $\downarrow$ \\ (ROU)}  \\
        \midrule
        INADE & 21.24 & 0.28 & 64.75 & 10.36 & 0.34 & 23.15 & 5.92 & 0.34 & 11.12 & 12.29 & 0.32 & 32.02 & 10.605 & 0.574 \\
        PZ20 & 6.27 & 0.61 & 25.76 & 3.77 & 0.67 & 10.74 & 2.43 & 0.67 & 6.74 & 4.12 & 0.65 & 14.04 & 20.449 & 0.234\\
        PMD & 3.35 & 0.47 & 5.77 & 2.34 & 0.53 & 2.98 & 1.49 & 0.56 & 1.78 & 2.39 & 0.52 & 3.46 & 17.638 & 0.176 \\
        LY21 & 3.41 & 0.58 & 9.02 & 2.36 & 0.63 & 4.47 & 1.61 & 0.63 & 2.66 & 2.45 & 0.61 & 5.29 & 19.366 & 0.151 \\
        CoCosNet & 4.95 & 0.57 & 15.56 & 3.53 & 0.61 & 6.21 & 1.96 & 0.60 & 3.18 & 3.49 & 0.59 & 8.10 & 18.361 & 0.249 \\
        UNITE & 3.07 & 0.76 & 6.27 & 2.10 & 0.81 & 2.32 & 1.40 & 0.81 & 1.26 & 2.18 & 0.79 & 3.19 & 22.272 & 0.096 \\
        CoCosNet v2 & 5.02 & 0.61 & 15.90 & 3.48 & 0.65 & 5.93 & 2.21 & 0.64 & 3.39 & 3.58 & 0.63 & 8.26 & 18.821 & 0.243 \\
        MCL-Net & 2.83 & 0.84 & 6.08 & 1.85 & 0.87 & 2.07 & 1.33 & 0.86 & 1.19 & 2.01 & 0.86 & 3.05 & 25.688 & 0.087 \\
        DynaST & 4.87 & 0.59 & 14.12 & 3.86 & 0.65 & 6.51 & 2.03 & 0.71 & 2.45 & 3.64 & 0.65 & 7.67 & 19.772 & 0.194 \\
        MATEBIT & 4.12 & 0.64 & 13.34 & 3.51 & 0.69 & 6.20 & 1.84 & 0.73 & 2.33 & 3.21 & 0.69 & 7.27 & 20.541 & 0.173 \\
        \midrule
        SIM-Net & \textbf{2.20} & 0.95 & \textbf{2.24} & \textbf{1.59} & 0.95 & \textbf{1.15} & 1.07 & 0.96 & 0.60 & 1.62 & 0.95 & \textbf{1.31} & 29.570 & 0.031 \\
        SIM-Net {\it w} DM & 2.23 & \textbf{0.97} & 2.29 & 1.61 &\textbf{0.96} & 1.18 & \textbf{1.02} & \textbf{0.97} & \textbf{0.54} & \textbf{1.63} & \textbf{0.97} & 1.34 & \textbf{30.124} & \textbf{0.028} \\
        \bottomrule
        \end{tabular}
    }
    \caption{Quantitative comparison in terms of Style Loss, SSIM, PSNR, and LPIPS. DM denotes as diffusion model. }
    \label{tab:overall}
\end{table*}

\noindent\textbf{Training dataset.}
Our training data is composed of 1,000 faces images from CelebAMask-HQ \cite{lee2020maskgan}, 227 horses images from Weizmann Horse Database \cite{borenstein2004combining}, and 696/573 mountains/buildings images from Intel Image Classification.
The size is 2,496, which is significantly smaller as compared with other GAN-based methods (\textit{e.g.}, 20,210 images from ADE20k \cite{zhou2017scene} used by \cite{zhang2020cross, zhou2021cocosnet, zhan2021unbalanced}). 

\noindent\textbf{Testing dataset.}
We collect 237 artistic images with 10 styles for evaluation, including painting styles of abstract (22 images, abbr. as {\textsc{Ab}}, sic passim), cubism (29 {\textsc{Cu}}), India (22 {\textsc{In}}), Chinese (21 {\textsc{Ch}}), expression (21 {\textsc{Ex}}), Japanese (32 {\textsc{Ja}}), modernism (21 {\textsc{Mo}}), impressionism (24 {\textsc{Im}}), photorealistic (24  {\textsc{Ph}}), and surrealism (21 {\textsc{Su}}).

\noindent\textbf{Evaluation Metrics}.
Since our method is not a GAN-based approach, and our training data and generated images are with quite different domains, 
metrics such as FID \cite{heusel2017gans} that aims to compute the distance between Gaussian fitted feature distributions of realistic and generated images, and SWD \cite{karras2018progressive} that attempts to measure the Wasserstein distance between the distribution of realistic images and generated images are not suitable for evaluation.
We present quantitative evaluation from several directions.
(1) Style Loss \cite{gatys2016image}, which has been used to measure the style similarity \cite{wu2021styleformer, wang2020collaborative}, is the mean-squared distance of features extracted from a pre-trained VGG model \cite{simonyan2014very}. 
(2) SSIM \cite{wang2003multiscale}, PSNR \cite{huynh2008scope} and LPIPS \cite{zhang2018unreasonable} are used to measure the image quality. 

\subsection{Overall Performance}
\noindent\textbf{Comparison with state-of-the-art.}
We compare qualitative and quantitative performance with state-of-the-art image manipulation methods, including two latent vector editing approaches INADE \cite{tan2021diverse} and swapping autoencoder \cite{park2020swapping} (abbr. as PZ20), two semantic correspondence approaches  PMD \cite{zhai2021Mutual} and learning to warp \cite{liu2021learning} (for general neural style transfer, abbr. as LY21), and exemplar-based image translation methods CoCosNet \cite{zhang2020cross}, UNITE \cite{zhan2021unbalanced}, CoCosNet v2 \cite{zhou2021cocosnet}, MCL-Net \cite{zhan2022marginal}, DynaST \cite{liu2022dynast} and MATEBIT \cite{jiang2023masked}.
Note that these methods as well as SIM-Net are trained on realistic images.
Furthermore, we propose an alternative version, SIM-Net w DM, which combines our method with a pretrained diffusion model.

\noindent\textbf{Quantitative evaluation.}
The quantitative comparison results are shown in Table \ref{tab:overall}. 
As can be observed from Table \ref{tab:overall}, comparison methods (\textit{e.g.}, INADE, UNITE) produce less competitive results with styles of {\textsc{Ab}}, {\textsc{Cu}}, and {\textsc{In}}, while achieving better results for those of {\textsc{Im}}, {\textsc{Ph}}, and {\textsc{Su}}.
Because these works are more likely to glean visual features of realistic training data and the former/latter styles are less/more realistic. 
Compared with existing methods, our model achieves competitive performance across all style images.

\begin{figure*}[!ht]
    \centering
    \includegraphics[width=0.98\textwidth]{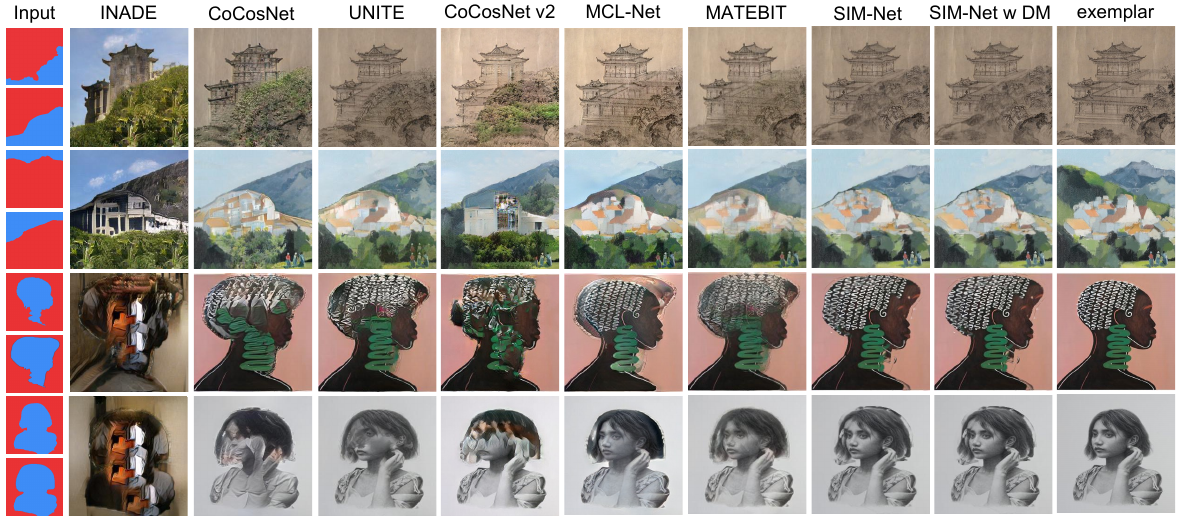}
    \caption{Visual qualitative comparison with state-of-the-art methods. 
    It can be seen that our method has no cross-domain artifacts and fine details without blurring. }
    \label{fig:all}
\end{figure*}

\noindent\textbf{Qualitative evaluation.}
Figure \ref{fig:all} illustrates images generated by different state-of-the-art methods.
Previous methods present local inconsistency and cross-domain artifacts due to the training data prior and space compression. 
In contrast, our results preserve consistent structure with edited mask, consistent style with exemplar image, and present consistent appearance with similar regions with the exemplar image. 

\begin{figure}[t]
    \centering
    \includegraphics[width=0.46\textwidth]{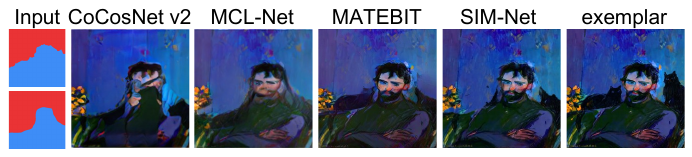}
    \caption{Qualitative comparison of our method and state-of-the-art methods in terms of high-resolution. }
    \label{fig:res}
\end{figure}

\begin{table}[t]
    \centering
    \resizebox{0.48\textwidth}{!}
    {
        \begin{tabular} {l|cc|c|cc}
        \bottomrule
        Method & Color $\uparrow$ & Texture $\uparrow$  & fake detection $\downarrow$ & Time $\downarrow$ & Memory $\downarrow$\\
        \midrule
        CoCosNet & 0.752 & 0.536 & 0.42 & 0.716 & 162.9 \\
        CoCosNet v2& 0.795 & 0.616 & 0.38 & 3.248 & \textbf{59.3} \\
        MCL-Net & 0.884 & 0.758 & 0.25 & 0.574 & 215 \\
        MATEBIT & 0.847 & 0.771 & 0.31 & 0.301 & 114.7 \\
        \midrule
        SIM-Net &\textbf{0.925} &\textbf{0.814} & \textbf{0.17} & \textbf{0.126} & 89.64\\
        \bottomrule
        \end{tabular}
    }
    \caption{Quantitative evaluation in terms of style relevance (color and texture), fake detection, time, and memory. }
    \label{table:style_relevance}
\end{table}

\noindent\textbf{Inference Speed and Computational Efficiency.}
To prove the advantages of our method's low computation and full precision, we choose to conduct comparative experiments with state-of-the-art methods on 20 high-resolution images.
For a fair comparison, we choose to use the results without a diffusion module.
As shown in Table \ref{table:style_relevance}, our method achieves a balance of full resolution and low computation, and maintains fast inference speed.
As illustrated in Figure \ref{fig:res}, our method achieves precise manipulation and guarantees high resolution.
Besides, when there is a significant difference between the edited mask and the input, our method can address it through multi-step operations with fast response.

\begin{figure}[t]
    \centering
    \includegraphics[width=0.48\textwidth]{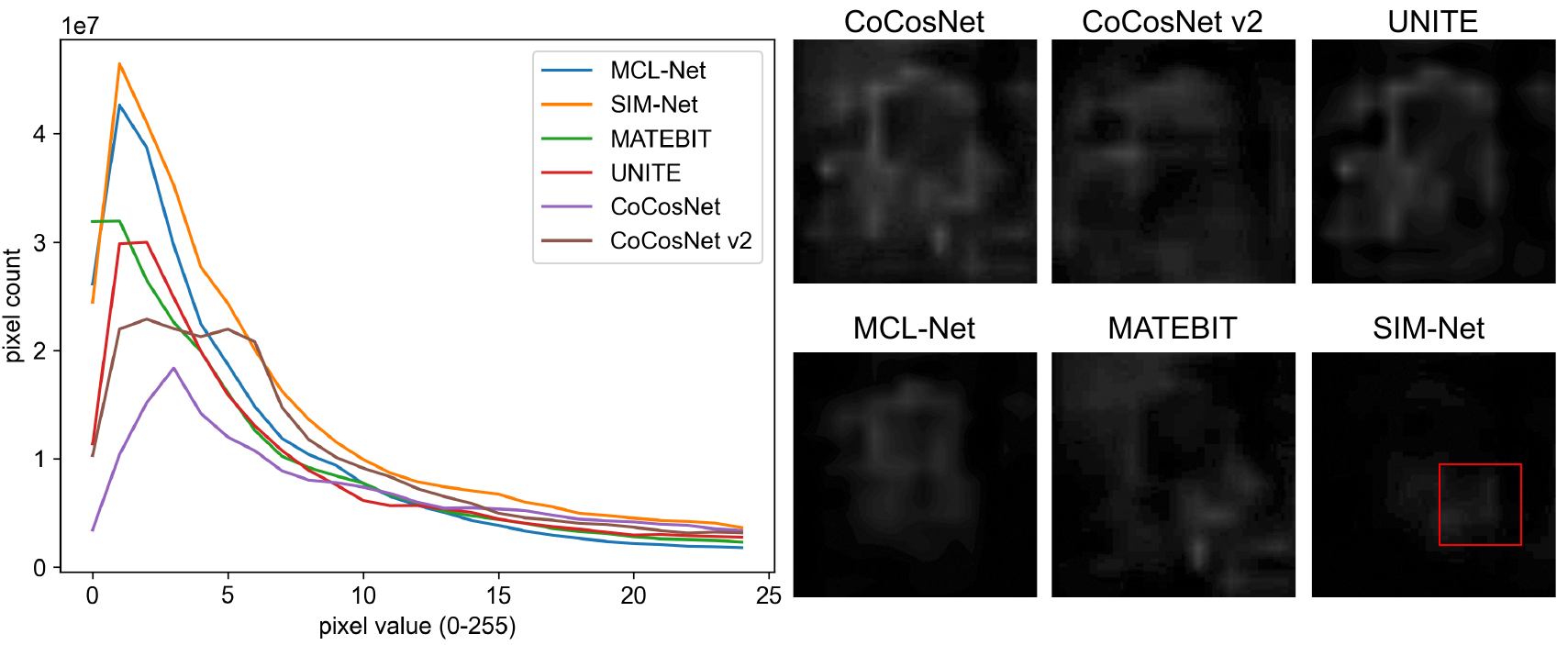}
    \caption{Qualitative and quantitative comparison for manipulation artifacts (row 1 in Figure \ref{fig:all}). The horizontal axis represents pixel values (0-255), and the vertical axis represents the pixel count. The red box has slight tampering marks.}
    \label{fig:fd}
\end{figure}

\noindent\textbf{Effectiveness of Eliminating Cross-Domain Artifacts.}
The previous style loss demonstrated the overall difference between images, and we further demonstrate that our method does not introduce cross-domain artifacts. 
Specifically, the style relevance is proved through the three dimensions of color, texture and fake detection. 
We compare the cosine similarities between low level features to measure color and texture relevance and use mvssnet++ \cite{chen2021image, dong2022mvss} for fake detection with a threshold of $0.04$.
Table \ref{table:style_relevance} shows that our method has clear advantages over existing methods.
Figure \ref{fig:fd} illustrates that our method has relatively fewer artifacts of manipulation.

\subsection{Ablation Study}

Table \ref{table:ablation} shows our method (full model) demonstrates best performance.
Specifically, 
(1) remove $\mathcal{L}_{\rm bound}$ to validate structural consistency (see SSIM);
(2) remove $\mathcal{L}_{\rm context}$ to validate style minimization (see Style Loss);
(3) remove $\mathcal{L}_{\rm cyc}$ to validate the importance for the mask-based correspondence network;
(4) replace semantic-free mask with semantic mask to explore the impact of semantic information;
(5) remove style self-supervision strategy to validate the necessity on generative quality, which tends to produce results with blurry artifacts;
(6) replace our generation network with GANs to validate its effectiveness.
  
\begin{table}[htbp]
    \centering
    \resizebox{0.48\textwidth}{!}
    {
        \begin{tabular} {l|cccc}
        \bottomrule
        & Style Loss $\downarrow$ & SSIM $\uparrow$  & PSNR $\uparrow$ & LPIPS $\downarrow$\\
        \midrule
        {\it w/o} $\mathcal{L}_{\rm bound}$ & 1.78 & 0.79 & 28.114 & 0.057 \\
        {\it w/o} $\mathcal{L}_{\rm context}$ & 2.24 & 0.92 & 25.126 & 0.174 \\
        {\it w/o} $\mathcal{L}_{\rm cyc}$ & 1.92 & 0.72 & 26.358 & 0.122 \\
        \midrule
        {\it w} semantic mask & \textbf{1.09} & 0.96 & 29.963 & 0.033 \\
        {\it w/o} Style Self-Supervision & 10.16 & 0.76 & 22.440 & 0.184 \\
        {\it w} generation model & 8.19 & 0.62 & 20.374 & 0.151 \\
        \midrule
        Full model: SIM-Net & 1.34 & \textbf{0.97} & \textbf{30.124} & \textbf{0.028} \\
        \bottomrule
        \end{tabular}
    }
    \caption{Ablation study of SIM-Net. }
    \label{table:ablation}
\end{table}

\begin{figure}[t]
    \centering
    \includegraphics[width=0.48\textwidth]{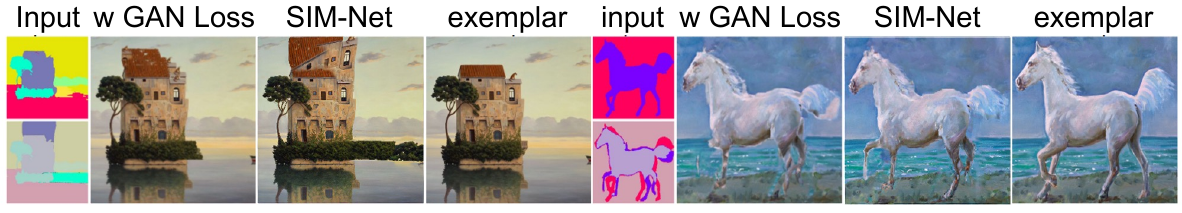}
    \caption{Introduce GAN loss which makes the representation effect insensitive to locality. Note that conditional semantic masks are enhanced to highlight their differences for better visualization.}
    \label{fig:gan}
\end{figure}

\noindent\textbf{Unnecessity of GAN Loss.}
To prove the Unnecessity of GAN Loss, we have tried using adversarial loss by adding an additional discriminator (the same network architecture in \cite{zhan2022marginal}). 
As can be observed from Figure \ref{fig:gan}, although the cross-domain artifacts are suppressed by our method, the results are insensitive to local manipulation.

\section{Conclusion}
This paper presents SIM-Net, a novel zero-shot image manipulation framework. 
The proposed scheme is capable of handling diverse styles of artistic images without relying on large-scale image training. 
Through quantitative and quality experiments with state-of-the-art methods, our approach demonstrates effective image processing capabilities.

\noindent\textbf{Limitations.} 
Our method is less creative than generative models, and is suitable for precisely controlled scenarios.
Our method affects artistic aesthetics, such as brush strokes, which is an advanced issue in the application of AI to art, requiring prior knowledge from domain experts \cite{he2018chipgan}.
This remains an open challenge for research.

\section{Acknowledgements}
This work is supported by the National Natural Science Foundation of China (No.U20A20220), the grants from Key R\&D Program of Zhejiang (no. 2022C01048), Key Program of National Natural Science Foundation of China (62334014).

\bibliography{aaai24}

\begin{thebibliography}{54}
\providecommand{\natexlab}[1]{#1}

\bibitem[{Bai, Nakashima, and Garcia(2021)}]{bai2021explain}
Bai, Z.; Nakashima, Y.; and Garcia, N. 2021.
\newblock Explain Me the Painting: Multi-Topic Knowledgeable Art Description Generation.
\newblock In \emph{Int. Conf. Comput. Vis.}

\bibitem[{Borenstein, Sharon, and Ullman(2004)}]{borenstein2004combining}
Borenstein, E.; Sharon, E.; and Ullman, S. 2004.
\newblock Combining top-down and bottom-up segmentation.
\newblock In \emph{IEEE Conf. Comput. Vis. Pattern Recog. Worksh.}

\bibitem[{Chen and Yang(2019)}]{chen2019recognizing}
Chen, L.; and Yang, J. 2019.
\newblock Recognizing the style of visual arts via adaptive cross-layer correlation.
\newblock In \emph{ACM Int. Conf. Multimedia}.

\bibitem[{Chen et~al.(2020)Chen, Su, Gao, Xia, and Fu}]{chen2020deepfacedrawing}
Chen, S.-Y.; Su, W.; Gao, L.; Xia, S.; and Fu, H. 2020.
\newblock DeepFaceDrawing: Deep generation of face images from sketches.
\newblock \emph{ACM Trans. Graph.}

\bibitem[{Chen et~al.(2021)Chen, Dong, Ji, Cao, and Li}]{chen2021image}
Chen, X.; Dong, C.; Ji, J.; Cao, J.; and Li, X. 2021.
\newblock Image manipulation detection by multi-view multi-scale supervision.
\newblock In \emph{Int. Conf. Comput. Vis.}

\bibitem[{Cheng, Schwing, and Kirillov(2021)}]{cheng2021per}
Cheng, B.; Schwing, A.; and Kirillov, A. 2021.
\newblock Per-pixel classification is not all you need for semantic segmentation.
\newblock \emph{Adv. Neural Inform. Process. Syst.}

\bibitem[{Dogan, Gu, and Timofte(2019)}]{dogan2019exemplar}
Dogan, B.; Gu, S.; and Timofte, R. 2019.
\newblock Exemplar guided face image super-resolution without facial landmarks.
\newblock In \emph{IEEE Conf. Comput. Vis. Pattern Recog. Worksh.}

\bibitem[{Dong et~al.(2022)Dong, Chen, Hu, Cao, and Li}]{dong2022mvss}
Dong, C.; Chen, X.; Hu, R.; Cao, J.; and Li, X. 2022.
\newblock Mvss-net: Multi-view multi-scale supervised networks for image manipulation detection.
\newblock \emph{IEEE Trans. Pattern Anal. Mach. Intell.}

\bibitem[{Fan et~al.(2022)Fan, Huang, Zheng, Liu, Qin, and Van~Gool}]{fan2022facial}
Fan, D.-P.; Huang, Z.; Zheng, P.; Liu, H.; Qin, X.; and Van~Gool, L. 2022.
\newblock Facial-sketch synthesis: a new challenge.
\newblock \emph{Machine Intelligence Research}.

\bibitem[{Gatys, Ecker, and Bethge(2016)}]{gatys2016image}
Gatys, L.~A.; Ecker, A.~S.; and Bethge, M. 2016.
\newblock Image style transfer using convolutional neural networks.
\newblock In \emph{IEEE Conf. Comput. Vis. Pattern Recog.}

\bibitem[{He et~al.(2018)He, Gao, Ma, Shi, and Duan}]{he2018chipgan}
He, B.; Gao, F.; Ma, D.; Shi, B.; and Duan, L.-Y. 2018.
\newblock Chipgan: A generative adversarial network for chinese ink wash painting style transfer.
\newblock In \emph{ACM Int. Conf. Multimedia}.

\bibitem[{He et~al.(2017)He, Gkioxari, Doll{\'a}r, and Girshick}]{he2017mask}
He, K.; Gkioxari, G.; Doll{\'a}r, P.; and Girshick, R. 2017.
\newblock Mask r-cnn.
\newblock In \emph{Int. Conf. Comput. Vis.}

\bibitem[{Heusel et~al.(2017)Heusel, Ramsauer, Unterthiner, Nessler, and Hochreiter}]{heusel2017gans}
Heusel, M.; Ramsauer, H.; Unterthiner, T.; Nessler, B.; and Hochreiter, S. 2017.
\newblock Gans trained by a two time-scale update rule converge to a local nash equilibrium.
\newblock \emph{Adv. Neural Inform. Process. Syst.}

\bibitem[{Huynh-Thu and Ghanbari(2008)}]{huynh2008scope}
Huynh-Thu, Q.; and Ghanbari, M. 2008.
\newblock Scope of validity of PSNR in image/video quality assessment.
\newblock \emph{Electronics letters}.

\bibitem[{Isola et~al.(2017)Isola, Zhu, Zhou, and Efros}]{isola2017image}
Isola, P.; Zhu, J.-Y.; Zhou, T.; and Efros, A.~A. 2017.
\newblock Image-to-image translation with conditional adversarial networks.
\newblock In \emph{IEEE Conf. Comput. Vis. Pattern Recog.}

\bibitem[{Jiang et~al.(2023)Jiang, Gao, Ma, Lin, Wang, and Xu}]{jiang2023masked}
Jiang, C.; Gao, F.; Ma, B.; Lin, Y.; Wang, N.; and Xu, G. 2023.
\newblock Masked and Adaptive Transformer for Exemplar Based Image Translation.
\newblock In \emph{IEEE Conf. Comput. Vis. Pattern Recog.}

\bibitem[{Karras et~al.(2018)Karras, Aila, Laine, and Lehtinen}]{karras2018progressive}
Karras, T.; Aila, T.; Laine, S.; and Lehtinen, J. 2018.
\newblock Progressive Growing of GANs for Improved Quality, Stability, and Variation.
\newblock In \emph{Int. Conf. Learn. Represent.}

\bibitem[{Lee et~al.(2020)Lee, Liu, Wu, and Luo}]{lee2020maskgan}
Lee, C.-H.; Liu, Z.; Wu, L.; and Luo, P. 2020.
\newblock Maskgan: Towards diverse and interactive facial image manipulation.
\newblock In \emph{IEEE Conf. Comput. Vis. Pattern Recog.}

\bibitem[{Li and Pun(2023)}]{li2023cee}
Li, H.; and Pun, C.-M. 2023.
\newblock CEE-Net: complementary end-to-end network for 3D human pose generation and estimation.
\newblock In \emph{AAAI}.

\bibitem[{Li et~al.(2021)Li, Fan, Yang, Luo, Cheng, and Liu}]{zhai2021Mutual}
Li, X.; Fan, D.-P.; Yang, F.; Luo, A.; Cheng, H.; and Liu, Z. 2021.
\newblock Probabilistic Model Distillation for Semantic Correspondence.
\newblock In \emph{IEEE Conf. Comput. Vis. Pattern Recog.}

\bibitem[{Li, Zuo, and Loy(2023)}]{li2023learning}
Li, X.; Zuo, W.; and Loy, C.~C. 2023.
\newblock Learning Generative Structure Prior for Blind Text Image Super-Resolution.
\newblock In \emph{IEEE Conf. Comput. Vis. Pattern Recog.}

\bibitem[{Liang, Zeng, and Zhang(2021)}]{liang2021high}
Liang, J.; Zeng, H.; and Zhang, L. 2021.
\newblock High-resolution photorealistic image translation in real-time: A laplacian pyramid translation network.
\newblock In \emph{IEEE Conf. Comput. Vis. Pattern Recog.}

\bibitem[{Liu, Breuel, and Kautz(2017)}]{liu2017unsupervised}
Liu, M.-Y.; Breuel, T.; and Kautz, J. 2017.
\newblock Unsupervised image-to-image translation networks.
\newblock \emph{Adv. Neural Inform. Process. Syst.}

\bibitem[{Liu et~al.(2022)Liu, Ye, Ren, and Wang}]{liu2022dynast}
Liu, S.; Ye, J.; Ren, S.; and Wang, X. 2022.
\newblock Dynast: Dynamic sparse transformer for exemplar-guided image generation.
\newblock In \emph{Eur. Conf. Comput. Vis.}

\bibitem[{Liu, Yang, and Hall(2021)}]{liu2021learning}
Liu, X.-C.; Yang, Y.-L.; and Hall, P. 2021.
\newblock Learning To Warp for Style Transfer.
\newblock In \emph{IEEE Conf. Comput. Vis. Pattern Recog.}

\bibitem[{Liu et~al.(2015)Liu, Luo, Wang, and Tang}]{liu2015deep}
Liu, Z.; Luo, P.; Wang, X.; and Tang, X. 2015.
\newblock Deep learning face attributes in the wild.
\newblock In \emph{Int. Conf. Comput. Vis.}

\bibitem[{Lugmayr et~al.(2022)Lugmayr, Danelljan, Romero, Yu, Timofte, and Van~Gool}]{lugmayr2022repaint}
Lugmayr, A.; Danelljan, M.; Romero, A.; Yu, F.; Timofte, R.; and Van~Gool, L. 2022.
\newblock Repaint: Inpainting using denoising diffusion probabilistic models.
\newblock In \emph{IEEE Conf. Comput. Vis. Pattern Recog.}

\bibitem[{Ma et~al.(2017{\natexlab{a}})Ma, Gao, Bai, Lou, Wang, Huang, and Duan}]{ma2017part}
Ma, D.; Gao, F.; Bai, Y.; Lou, Y.; Wang, S.; Huang, T.; and Duan, L.-Y. 2017{\natexlab{a}}.
\newblock From part to whole: who is behind the painting?
\newblock In \emph{ACM Int. Conf. Multimedia}.

\bibitem[{Ma et~al.(2017{\natexlab{b}})Ma, Jia, Sun, Schiele, Tuytelaars, and Van~Gool}]{ma2017pose}
Ma, L.; Jia, X.; Sun, Q.; Schiele, B.; Tuytelaars, T.; and Van~Gool, L. 2017{\natexlab{b}}.
\newblock Pose guided person image generation.
\newblock \emph{Adv. Neural Inform. Process. Syst.}

\bibitem[{Mechrez, Talmi, and Zelnik-Manor(2018)}]{mechrez2018contextual}
Mechrez, R.; Talmi, I.; and Zelnik-Manor, L. 2018.
\newblock The contextual loss for image transformation with non-aligned data.
\newblock In \emph{Eur. Conf. Comput. Vis.}

\bibitem[{Men et~al.(2020)Men, Mao, Jiang, Ma, and Lian}]{men2020controllable}
Men, Y.; Mao, Y.; Jiang, Y.; Ma, W.-Y.; and Lian, Z. 2020.
\newblock Controllable person image synthesis with attribute-decomposed gan.
\newblock In \emph{IEEE Conf. Comput. Vis. Pattern Recog.}

\bibitem[{Nazeri et~al.(2019)Nazeri, Ng, Joseph, Qureshi, and Ebrahimi}]{nazeri2019edgeconnect}
Nazeri, K.; Ng, E.; Joseph, T.; Qureshi, F.; and Ebrahimi, M. 2019.
\newblock Edgeconnect: Structure guided image inpainting using edge prediction.
\newblock In \emph{Int. Conf. Comput. Vis. Worksh.}

\bibitem[{Pan et~al.(2023)Pan, Tewari, Leimk{\"u}hler, Liu, Meka, and Theobalt}]{pan2023drag}
Pan, X.; Tewari, A.; Leimk{\"u}hler, T.; Liu, L.; Meka, A.; and Theobalt, C. 2023.
\newblock Drag your gan: Interactive point-based manipulation on the generative image manifold.
\newblock In \emph{ACM SIGGRAPH 2023 Conference Proceedings}.

\bibitem[{Park et~al.(2019)Park, Liu, Wang, and Zhu}]{park2019semantic}
Park, T.; Liu, M.-Y.; Wang, T.-C.; and Zhu, J.-Y. 2019.
\newblock Semantic image synthesis with spatially-adaptive normalization.
\newblock In \emph{IEEE Conf. Comput. Vis. Pattern Recog.}

\bibitem[{Park et~al.(2020)Park, Zhu, Wang, Lu, Shechtman, Efros, and Zhang}]{park2020swapping}
Park, T.; Zhu, J.-Y.; Wang, O.; Lu, J.; Shechtman, E.; Efros, A.~A.; and Zhang, R. 2020.
\newblock Swapping Autoencoder for Deep Image Manipulation.
\newblock In \emph{Adv. Neural Inform. Process. Syst.}

\bibitem[{Siarohin et~al.(2019)Siarohin, Lathuili{\`e}re, Tulyakov, Ricci, and Sebe}]{siarohin2019first}
Siarohin, A.; Lathuili{\`e}re, S.; Tulyakov, S.; Ricci, E.; and Sebe, N. 2019.
\newblock First order motion model for image animation.
\newblock \emph{Adv. Neural Inform. Process. Syst.}

\bibitem[{Siarohin et~al.(2021)Siarohin, Woodford, Ren, Chai, and Tulyakov}]{siarohin2021motion}
Siarohin, A.; Woodford, O.~J.; Ren, J.; Chai, M.; and Tulyakov, S. 2021.
\newblock Motion representations for articulated animation.
\newblock In \emph{IEEE Conf. Comput. Vis. Pattern Recog.}

\bibitem[{Simonyan and Zisserman(2014)}]{simonyan2014very}
Simonyan, K.; and Zisserman, A. 2014.
\newblock Very deep convolutional networks for large-scale image recognition.
\newblock \emph{arXiv preprint arXiv:1409.1556}.

\bibitem[{Tan et~al.(2021)Tan, Chai, Chen, Liao, Chu, Liu, Hua, and Yu}]{tan2021diverse}
Tan, Z.; Chai, M.; Chen, D.; Liao, J.; Chu, Q.; Liu, B.; Hua, G.; and Yu, N. 2021.
\newblock Diverse Semantic Image Synthesis via Probability Distribution Modeling.
\newblock In \emph{IEEE Conf. Comput. Vis. Pattern Recog.}

\bibitem[{Wang et~al.(2020)Wang, Li, Wang, Hu, and Yang}]{wang2020collaborative}
Wang, H.; Li, Y.; Wang, Y.; Hu, H.; and Yang, M.-H. 2020.
\newblock Collaborative distillation for ultra-resolution universal style transfer.
\newblock In \emph{IEEE Conf. Comput. Vis. Pattern Recog.}

\bibitem[{Wang, Simoncelli, and Bovik(2003)}]{wang2003multiscale}
Wang, Z.; Simoncelli, E.~P.; and Bovik, A.~C. 2003.
\newblock Multiscale structural similarity for image quality assessment.
\newblock In \emph{The Thrity-Seventh Asilomar Conference on Signals, Systems \& Computers, 2003}.

\bibitem[{Wu et~al.(2021)Wu, Hu, Sheng, and Xu}]{wu2021styleformer}
Wu, X.; Hu, Z.; Sheng, L.; and Xu, D. 2021.
\newblock StyleFormer: Real-Time Arbitrary Style Transfer via Parametric Style Composition.
\newblock In \emph{Int. Conf. Comput. Vis.}

\bibitem[{Xu, Zhu, and Wang(2020)}]{xu2020generative}
Xu, S.; Zhu, Q.; and Wang, J. 2020.
\newblock Generative image completion with image-to-image translation.
\newblock \emph{Neural Computing and Applications}.

\bibitem[{Xu, Long, and Nie(2023)}]{xu2023learning}
Xu, W.; Long, C.; and Nie, Y. 2023.
\newblock Learning Dynamic Style Kernels for Artistic Style Transfer.
\newblock In \emph{IEEE Conf. Comput. Vis. Pattern Recog.}

\bibitem[{Zhan et~al.(2021)Zhan, Yu, Cui, Zhang, Lu, Pan, Zhang, Ma, Xie, and Miao}]{zhan2021unbalanced}
Zhan, F.; Yu, Y.; Cui, K.; Zhang, G.; Lu, S.; Pan, J.; Zhang, C.; Ma, F.; Xie, X.; and Miao, C. 2021.
\newblock Unbalanced Feature Transport for Exemplar-based Image Translation.
\newblock In \emph{IEEE Conf. Comput. Vis. Pattern Recog.}

\bibitem[{Zhan et~al.(2022)Zhan, Yu, Wu, Zhang, Lu, and Zhang}]{zhan2022marginal}
Zhan, F.; Yu, Y.; Wu, R.; Zhang, J.; Lu, S.; and Zhang, C. 2022.
\newblock Marginal contrastive correspondence for guided image generation.
\newblock In \emph{IEEE Conf. Comput. Vis. Pattern Recog.}

\bibitem[{Zhang, Zhan, and Chang(2021)}]{zhang2021deep}
Zhang, C.; Zhan, F.; and Chang, Y. 2021.
\newblock Deep monocular 3d human pose estimation via cascaded dimension-lifting.
\newblock \emph{arXiv preprint arXiv:2104.03520}.

\bibitem[{Zhang, Rao, and Agrawala(2023)}]{zhang2023adding}
Zhang, L.; Rao, A.; and Agrawala, M. 2023.
\newblock Adding conditional control to text-to-image diffusion models.
\newblock In \emph{Int. Conf. Comput. Vis.}

\bibitem[{Zhang et~al.(2020)Zhang, Zhang, Chen, Yuan, and Wen}]{zhang2020cross}
Zhang, P.; Zhang, B.; Chen, D.; Yuan, L.; and Wen, F. 2020.
\newblock Cross-domain correspondence learning for exemplar-based image translation.
\newblock In \emph{IEEE Conf. Comput. Vis. Pattern Recog.}

\bibitem[{Zhang et~al.(2018)Zhang, Isola, Efros, Shechtman, and Wang}]{zhang2018unreasonable}
Zhang, R.; Isola, P.; Efros, A.~A.; Shechtman, E.; and Wang, O. 2018.
\newblock The unreasonable effectiveness of deep features as a perceptual metric.
\newblock In \emph{IEEE Conf. Comput. Vis. Pattern Recog.}

\bibitem[{Zhang, Zheng, and Pan(2022)}]{zhang2022easypainter}
Zhang, Y.; Zheng, Q.; and Pan, G. 2022.
\newblock EasyPainter: Customizing Your Own Paintings.
\newblock In \emph{CAAI International Conference on Artificial Intelligence}.

\bibitem[{Zhou et~al.(2017)Zhou, Zhao, Puig, Fidler, Barriuso, and Torralba}]{zhou2017scene}
Zhou, B.; Zhao, H.; Puig, X.; Fidler, S.; Barriuso, A.; and Torralba, A. 2017.
\newblock Scene parsing through ade20k dataset.
\newblock In \emph{IEEE Conf. Comput. Vis. Pattern Recog.}

\bibitem[{Zhou et~al.(2021)Zhou, Zhang, Zhang, Zhang, Bao, Chen, Zhang, and Wen}]{zhou2021cocosnet}
Zhou, X.; Zhang, B.; Zhang, T.; Zhang, P.; Bao, J.; Chen, D.; Zhang, Z.; and Wen, F. 2021.
\newblock CoCosNet v2: Full-Resolution Correspondence Learning for Image Translation.
\newblock In \emph{IEEE Conf. Comput. Vis. Pattern Recog.}

\bibitem[{Zhu et~al.(2017)Zhu, Park, Isola, and Efros}]{zhu2017unpaired}
Zhu, J.-Y.; Park, T.; Isola, P.; and Efros, A.~A. 2017.
\newblock Unpaired image-to-image translation using cycle-consistent adversarial networks.
\newblock In \emph{Int. Conf. Comput. Vis.}

\end{thebibliography}

\end{document}